\documentclass[11pt]{article}

\usepackage[final]{acl}

\usepackage{times}
\usepackage{latexsym}
\usepackage{amsmath}
\usepackage{tabularx}
\usepackage{booktabs}
\usepackage{multirow}
\usepackage{amssymb}
\usepackage{pifont}
\newcommand{\tick}{\ding{52}}
\newcommand{\cross}{\ding{56}}
\usepackage{makecell}
\usepackage{colortbl}
\usepackage[table]{xcolor}
\usepackage{adjustbox}
\usepackage{colortbl}

\usepackage[T1]{fontenc}

\usepackage[utf8]{inputenc}

\usepackage{microtype}

\usepackage{inconsolata}
\usepackage{tcolorbox}

\usepackage{graphicx}
\usepackage{fontawesome5}

\title{The Role of Ambiguity in Error Prediction via Uncertainty Quantification}

\author{
    Ieva Raminta Staliūnaitė\textsuperscript{\faGraduationCap, \faRobot} \quad 
    James Bishop\textsuperscript{\faRobot} \quad 
    Andreas Vlachos\textsuperscript{\faGraduationCap} \\
    \textsuperscript{\faGraduationCap}University of Cambridge \quad 
    \textsuperscript{\faRobot}The Alan Turing Institute \\
    \texttt{\{irs38, av308\}@cam.ac.uk} \quad \texttt{jbishop@turing.ac.uk}
}

\begin{document}
\maketitle

\begin{abstract}
The task of Error Prediction, namely predicting whether a model output is correct,
is commonly tackled with Uncertainty Quantification (UQ).
However,
while uncertainty metrics 
capture when models lack knowledge or capacity to 
make a prediction, they also reflect aleatoric uncertainty, which is inherent in the model input and context. 
This paper presents a method for improving error prediction for Large Language Models (LLMs), by disentangling input ambiguity from UQ signal.
We conduct experiments on the task of Question Answering (QA)  with six UQ metrics and show that UQ metrics are more predictive of errors on unambiguous instances than on questions with multiple plausible answers.
We use Gated Experts and Selective Prediction to incorporate gold and predicted ambiguity labels into the error prediction pipeline. 
We find that ambiguity information improves error prediction scores across 
model families, training and evaluation paradigms, datasets (including allegedly unambiguous ones), and sources of aleatoric uncertainty, yielding improvements of over 10 points of PRR for individual UQ metrics on standard datasets. 
\end{abstract}

\section{Introduction}

Predicting whether a model will correctly answer a question is important for establishing model reliability and reducing risks associated with incorrect model outputs. 
The task of error prediction is usually performed by leveraging model uncertainty,
yielding successful results \citep{kadavath2022language, lin2023generating, manakul-etal-2023-selfcheckgpt, farquhar2024detecting}. 
However, uncertainty comprises both aleatoric  and epistemic uncertainty. Aleatoric uncertainty is inherent to the data due to ambiguity, conflicting inputs or noise, and is therefore irreducible; whereas epistemic uncertainty, which stems from the model's lack of capacity or knowledge, is reducible \citep{kendall2017uncertainties, hullermeier2021aleatoric}. 

\begin{figure}[t]
    \centering
    \includegraphics[width=\linewidth]{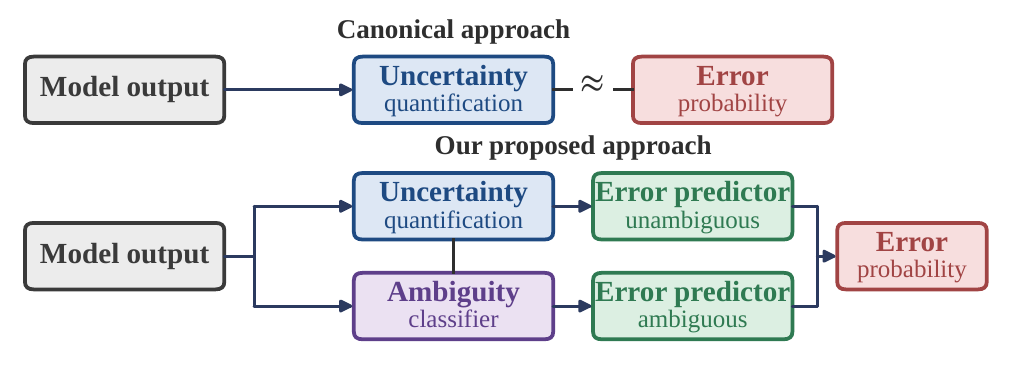}
    \caption{The established error prediction method directly using model features such as uncertainty metrics (top) and our proposed framework, predicting ambiguity and using it in a gated experts model for improving error prediction scores (bottom).
     }
    \label{fig:figure1}
\end{figure}

Ambiguity may be present at every step of the LLM pipeline.
It can arise from conflicting information in the training data, a user asking an underspecified question, or contradictions in the context. 
Question Answering (QA) offers an experimental setup where ambiguity occurs naturally. 
To illustrate this, consider the following 
example from AmbigQA~\citep{min-etal-2020-ambigqa}: \textit{``What is it called when you mix up the letters of a word?''} Valid answers include \textit{`dyslexia'} (referring to a learning disability) or \textit{`anagram'} (referring to an orthographic phenomenon). 
A model processing this question may be legitimately uncertain about the answer, given that there are at least two plausible interpretations. 
In such a case the canonical approach to error prediction (see top of Figure~\ref{fig:figure1}) would incorrectly predict high probability of error due to high model uncertainty, as it does not discriminate aleatoric and epistemic uncertainty. 

Some work has shown that UQ metrics capture both aleatoric and epistemic uncertainty, and the entanglement of the two makes UQ metrics less reliable for error prediction in ambiguous settings.  
\citet{baan-etal-2024-interpreting} posit that standard predictive distributions inherently conflate epistemic uncertainty reflecting a model's lack of knowledge, and aleatoric uncertainty reflecting inherent input ambiguity and valid human label variation. 
\citet{tomov2025illusion} formally prove that consistency-based and ensemble-based UQ estimators are guaranteed to correlate with epistemic uncertainty only when aleatoric uncertainty is zero, and show empirically that their performance degrades to near-random on ambiguous QA datasets.

In this work we focus on
improving error prediction by leveraging ambiguity. 
We implement two approaches for separating out the aleatoric uncertainty in UQ metrics.
First, a gated expert model uses the ambiguity signal as a routing mechanism, directing queries to separate, specialized sub-networks, one trained for ambiguous instances and one for unambiguous instances, to optimize error prediction for each specific domain independently.
Second, a selective prediction model uses the ambiguity signal to explicitly adjust the uncertainty score, selectively rejecting error-prone answers by penalizing or favoring samples based on their ambiguity. 

In our experiments we consider six Uncertainty Quantification (UQ) metrics, namely Maximum Softmax Probability (MSP) \citep{hendrycks2016baseline}, Semantic Entropy \citep{farquhar2024detecting},  Mutual Information via Iterative Prompting (MI) \citep{yadkori2024believe}, Shifting Attention to Relevance (SAR) \citep{duan2024shifting}, Semantic Energy \citep{ma2025semantic} and CoCoA \citep{vashurinCoCoA}.
We carry out experiments on three datasets: AmbigQA~\citep{min-etal-2020-ambigqa}, NATCONFQA (NCQA)~\citep{nachshoni-etal-2025-consensus}, and TriviaQA~\citep{joshi-etal-2017-triviaqa}.
We show that ambiguity is useful in predicting errors on both ambiguous and unambiguous QA datasets, for various model families, including models that have been specifically finetuned to represent human label variation, for all uncertainty metrics considered, including their combinations.\footnote{The code will be released on \url{https://github.com/ieva-raminta/error_or_ambiguity}} 

\section{Related Work}
\label{sec:rel}

Research has shown that UQ metrics encode more than whether the model is likely to be wrong. This distinction has been demonstrated to be relevant in Hate Speech Detection and Sentiment Analysis \citep{davani2022dealing}, Word Sense Disambiguation \citep{liu2023ambiguity}, Natural Language Inference \citep{staliunaite-vlachos-2025-uncertain}, and Machine Translation \citep{staliunaite-etal-2026-uncertainty}. However, the question of whether explicitly modeling ambiguity can improve error prediction, rather than task accuracy, remains open. 
Most related to our work, \citet{cole2023selectively} show that sampling-based confidence scores are better calibrated than likelihood on ambiguous questions, and propose a disambiguate-then-answer paradigm to handle denotational uncertainty. They also attempt to predict ambiguity from the model's own outputs but report that none of their methods exceed chance by a meaningful margin, and consequently do not incorporate predicted ambiguity as an explicit signal in the abstention decision.
\citet{10.5555/3692070.3692835} and \citet{walha2026fine} similarly decompose predictive uncertainty into aleatoric and epistemic components via input clarification ensembling and a spectral kernel-based approach respectively, but target uncertainty estimation itself rather than using the decomposed signal to supervise error prediction.

LLMs have been shown to be overconfident in their most likely prediction \citep{tian-etal-2023-just}, and often confident in incorrect answers \citep{simhi2025trust}. This is particularly relevant when multiple answers are equally plausible. Some work has shown that modeling ambiguity or subjectivity boosts task performance \citep{uma2021learning, plank-2022-problem}. Recent finetuning methods address this more directly, namely \citet{sorensen2025spectrum} introduce Spectrum Tuning to enhance distributional coverage, and \citet{zhang20252} show that training on distributions or multi-sample prompting mitigates mode collapse. 
Furthermore, \citet{yang2025maqa} demonstrate that while the presence of inherent data uncertainty in multi-answer QA setups significantly degrades standard UQ metrics like max logit or verbalized confidence, entropy and response consistency remain robust estimators of model uncertainty.
\citet{yadkori2024believe} propose a UQ metric based on iterative prompting that explicitly targets epistemic uncertainty by measuring sensitivity to prior context. It requires multiple sequential prompting rounds and assumes aleatoric uncertainty manifests as context-insensitivity. None of these methods ask whether a model that better represents distributional uncertainty also becomes easier to supervise for error prediction.

Prior work has shown that internal model representations encode strong signals of output reliability and ambiguity.
Namely, \citet{Chen2024INSIDELI} leverage eigenvalues of response embedding covariance to detect hallucinations, \citet{Orgad2024LLMsKM} show that probing exact-answer tokens reveals fine-grained error-type information, and \citet{NEURIPS2023_81b83900} identify and steer truthfulness-related attention heads at inference time.
Furthermore, \citet{vashurinCoCoA} train a lightweight model on middle-layer embeddings to predict consistency-based uncertainty scores without repeated sampling. \citet{10.5555/3692070.3692092} train linear probes on internal embeddings to predict token-level confidence, finding middle layers most predictive. \citet{ch-wang-etal-2024-androids} show hallucination propensity can be detected from hidden states alone. 
We note that \citet{zhang-etal-2025-sparse-neurons} address a different task of classifying original questions against explicitly disambiguated counterparts, introducing systematic surface differences. Their signal peaks in early lexically-focused layers \citep{jin-etal-2025-exploring}, consistent with probes capturing surface form rather than semantic ambiguity.
This motivates including middle-layer representations as ambiguity classification features, where the signal of interest is semantic rather than surface-level.

Finally, \citet{huang2026richest} show that many TriviaQA questions suffer from underspecification and that detecting and rewriting them improves answering accuracy. \citet{zhang20252} similarly show that standard single-answer benchmarks admit multiple valid responses, and introduce a reinforcement learning framework mining alternative answers to improve model performance. It is unknown whether the latent ambiguity in such datasets also undermines the ability of UQ metrics to predict errors on them.

\section{Method}
\label{sec:method}

\begin{figure}
    \centering
    \includegraphics[width=.9\linewidth]{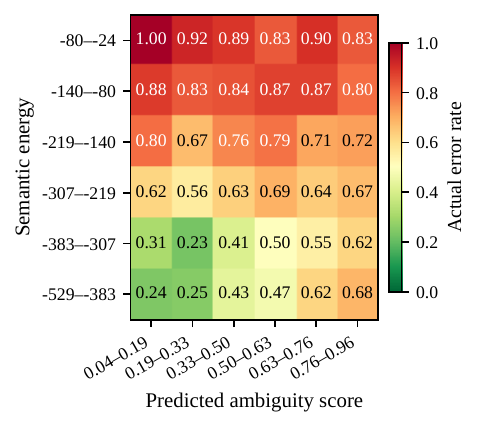}
    \caption{Heatmap illustrating the relationship between UQ (Semantic Entropy) and predicted ambiguity on the one hand, and the error rate on the other. Ambiguity scores are predicted by the model described in Section~\ref{sub:prederrors}; the same pattern holds with gold ambiguity labels, confirming it reflects a genuine data property rather than a modelling artefact.}
    \label{fig:heatmap}
\end{figure}

This section describes the approach proposed in this paper, namely extending the standard approach of using UQ metrics directly as error predictors (which we define formally as our baseline in Section~\ref{sub:errpredbase}, and to which we compare throughout).
The injection of ambiguity labels is motivated by Figure~\ref{fig:heatmap}, which illustrates that when there is little ambiguity (left hand-side of the figure), low uncertainty indicates low error likelihood and high uncertainty indicates high error rates, whereas when ambiguity is high (right) the error likelihood is similar for low and high uncertainty scores.
That is, UQ scores track error rates strongly on unambiguous inputs but only weakly on ambiguous ones, where error rates vary less with uncertainty. Conditioning on ambiguity lets the model trust UQ where it is informative and discount it where it is not.

\subsection{Features for Error Prediction and Ambiguity Classification}

We use three types of features to capture aleatoric and epistemic uncertainty signal, namely internal model representations, uncertainty quantification metrics, and semantic cluster statistics.

\subsubsection{Uncertainty Quantification}

We use UQ metrics as features in the ambiguity classifier (Section~\ref{sub:resultsambig}) and the error prediction models (Section~\ref{sub:prederrors}). We consider six methods, computed over a greedy decode and $K$ stochastic samples, that differ in how they aggregate confidence across samples.

\textbf{MSP} \citep{hendrycks2016baseline} uses one minus the highest sequence-level probability across samples as a simple confidence baseline.
\textbf{Semantic Entropy} \citep{farquhar2024detecting} clusters samples by bidirectional entailment and computes entropy over cluster probabilities, abstracting away lexical variation.
\textbf{Semantic Energy} \citep{ma2025semantic} replaces the cluster probability with a Boltzmann-style energy distribution over unnormalised logits, addressing cases where identical incorrect responses yield zero entropy.
\textbf{CoCoA} \citep{vashurinCoCoA} multiplies the greedy output's confidence by its mean pairwise dissimilarity to the samples, retaining the most-likely-output signal that Semantic Entropy discards.
\textbf{SAR} \citep{duan2024shifting} reweights token log-probabilities by their semantic relevance and adjusts sentence-level uncertainty by cross-sample similarity.
\textbf{MI} \citep{yadkori2024believe} targets epistemic uncertainty by computing the KL divergence between the empirical joint over responses from iterative prompting and the product of marginals.
\subsubsection{Semantic Cluster Features}

For ambiguity prediction, we additionally include semantic cluster properties of sampled responses. 
The cluster features include the number of clusters, cluster assignment entropy, effective number of answers, maximum cluster probability, the probability mass of the top two clusters, and the probability gap between the top two clusters.

To assess which features are most informative for ambiguity classification,
we measure the AUROC of each scalar feature as a standalone binary classifier
of the ambiguity label (ambiguous vs. unambiguous), presented in Table~\ref{tab:feat_corr_llama}.
For high-dimensional representation features (last-layer embeddings and
mid-layer residuals), we report the mean AUROC from 5-fold cross-validated
logistic regression after PCA reduction to 128 components.\footnote{The results for all models are presented in Appendix~\ref{app:correlation}, showing the same pattern.}

Uncertainty features show consistent positive correlation with ambiguity across datasets and models, with entropy-based and cluster-diversity features showing stronger correlations than concentration-based features such as max cluster probability. Representation features show stronger correlations than scalar features, particularly on AmbigQA. These results motivate their use as features in the ambiguity classifier described in Section~\ref{sub:prederrors}.

\subsubsection{Internal Representations}

Motivated by the body of work discussed in Section~\ref{sec:rel}, we incorporate two representation features: the final-layer hidden state embedding and a mid-layer residual from the last generated token, providing complementary views of the model's internal state at generation time.

\begin{table}[t]
\centering
\resizebox{\columnwidth}{!}{%
\setlength{\tabcolsep}{1pt}
\tiny
\begin{tabular}{l|cc|cccccc|cccccc}
\toprule
& \multicolumn{2}{c|}{\textbf{Internal}} & \multicolumn{6}{c|}{\textbf{UQ}} & \multicolumn{6}{c}{\textbf{Semantic Cluster}} \\
\textbf{Data} & \textbf{Emb} & \textbf{Res} & \textbf{SE} & \textbf{SEng} & \textbf{MSP} & \textbf{SAR} & \textbf{MI} & \textbf{CoCoA} & \textbf{\#Cl} & \textbf{ClE} & \textbf{MxC} & \textbf{ENA} & \textbf{PG2} & \textbf{Top2} \\ \midrule
Ambig & .71 & .73 & .54 & .54 & .54 & .55 & .54 & .54 & .54 & .54 & .46 & .54 & .46 & .45 \\
NCQA  & .49 & .60 & .50 & .52 & .57 & .55 & .54 & .53 & .59 & .50 & .52 & .50 & .52 & .43 \\
Trivia & .54 & .56 & .56 & .57 & .56 & .57 & .56 & .57 & .56 & .56 & .44 & .56 & .45 & .44 \\ \bottomrule
\end{tabular}
}
\caption{AUROC of individual features for predicting ambiguity (Llama-3.1-8B, validation split). Scalar features evaluated directly; representation features use 5-fold CV logistic regression. Emb = last-token embedding; Res = layer-15 residual.
\#Cl = number of clusters; ClE = cluster entropy; MxC = max cluster probability; ENA = effective number of answers; PG2 = probability gap (top-2); Top2 = top-2 probability mass.}
\label{tab:feat_corr_llama}
\end{table}

\subsection{The Role of Ambiguity in Error Prediction via UQ}
\label{sub:prederrors}

Having established the features used for ambiguity classification and error prediction, 
we now describe the model architectures that disentangle the ambiguity signal in error prediction.
Both build on the baseline error prediction head (Section~\ref{sub:errpredbase}), which maps a UQ feature vector $\mathbf{x}_f$ to a scalar error logit $\hat{c}$; the architectures below differ only in how they incorporate the ambiguity signal.

The \textbf{Gated Experts} model routes instances to specialized expert networks for ambiguous ($E_a$) and unambiguous ($E_u$) queries based on ambiguity signal to mitigate feature interference (see bottom of Figure~\ref{fig:figure1} for illustration). The \textbf{Oracle Version} hard-routes using ground-truth ambiguity labels $a \in \{0, 1\}$ such that $\hat{c} = (1 - a)\, E_u(\mathbf{x}_f) + a\, E_a(\mathbf{x}_f)$. The \textbf{Latent Version} soft-routes using the pre-trained ambiguity model probability $\hat{a}$ such that $\hat{c} = (1 - \hat{a})\, E_u(\mathbf{x}_f) + \hat{a}\, E_a(\mathbf{x}_f)$.

The \textbf{Selective Prediction Framework} \citep{JMLR:v11:el-yaniv10a, geifman2017selective} computes a rejection score $R$ that penalizes the error logit $\hat{c}$ proportionally to ambiguity, scaled by $\lambda$, used to rank instances for abstention. Unlike \citet{cole2023selectively}, who use sampling-based confidence scores to threshold abstention, our selective prediction framework explicitly incorporates ambiguity as a penalty on the error score, allowing the model to abstain preferentially on instances that are both uncertain and ambiguous. The \textbf{Oracle Version} uses ground-truth ambiguity $a$ yielding $R = -\hat{c} + \lambda\, a$, whereas the \textbf{Latent Version} uses the pre-trained ambiguity model probability $\hat{a}$ yielding $R = -\hat{c} + \lambda\, \hat{a}$.

\textbf{The Ambiguity Model} is trained independently of the error prediction pipeline. It takes the same scalar UQ features $\mathbf{x}_f$ as input, processes them through an MLP to obtain a latent representation $\mathbf{z}_a = \text{MLP}_a(\mathbf{x}_f)$, and predicts the probability of instance-level semantic ambiguity via a linear classifier $\hat{a} = \sigma(\mathbf{w}^\top \mathbf{z}_a + b)$, optimized using binary cross-entropy against ground-truth ambiguity labels. The resulting ambiguity probabilities $\hat{a}$ are used as gating or penalty signals in the models described above.

\section{Experiments}

This section describes how we tested the contributions of ambiguity signal in predicting errors. 
We discuss the experimental details of running LLMs for inference on QA datasets, evaluated their QA performance and measured UQ scores, which features we then used to predict errors and how we evaluated the latter task. 

\subsection{Data}
\label{sub:data}

\begin{table}[t]
\centering
\small
\renewcommand{\arraystretch}{1.3}
\begin{tabularx}{\columnwidth}{@{}X@{}}
\toprule
\textbf{AmbigQA} \quad (Train: 10,036 / Dev: 2,002 / Test: -) \\
\textbf{Question:} The most common type of rock in Earth's crust is? \\
\textbf{Answers:} Mafic rocks; Granite \\
\textbf{Ambiguity Explanation:} Ambiguous referent; mafic rocks dominate the oceanic crust, while granite is the most common in the continental crust. \\
\midrule
\textbf{TriviaQA} \quad (Train: 61,888 / Dev: 7,993 / Test: -) \\
\textbf{Question:} Marie Curie's country of birth? \\
\textbf{Answers:} Poland \\
\textbf{Ambiguity Explanation:} N/A \\
\midrule
\textbf{NCQA} \quad (Train: 234 / Dev: 100 / Test: 100) \\
\textbf{Question:} Were the Middle Ages the Dark Ages? \\
\textbf{Answers:} Yes; No \\
\textbf{Ambiguity Explanation:} Competing evidence; Roman institutional collapse supports the ``darkness'' narrative, while Islamic and Carolingian peaks demonstrate intellectual progress. \\
\bottomrule
\end{tabularx}
\caption{Examples and sizes of the datasets. AmbigQA and TriviaQA do not release test sets with answers.}
\label{tab:dataset_examples}
\end{table}

The \textbf{AmbigQA} dataset \citep{min-etal-2020-ambigqa} formalizes ambiguity in open-domain QA by extending NQ~\citep{kwiatkowski-etal-2019-natural} with multiple plausible answer sets. The authors find that over 50\% of natural queries contain inherent ambiguity, such as conflicting entity references or temporal dependencies.

The \textbf{TriviaQA} dataset \citep{joshi-etal-2017-triviaqa} is a large-scale reading comprehension benchmark containing question-answer pairs authored by trivia enthusiasts. As discussed in Section~\ref{sec:rel}, while it is not officially an ambiguous question dataset, approximately 16\% of the instances in the dataset contain ambiguous questions. 
We extend the TriviaQA dataset with underspecification annotations following the protocol of \citet{huang2026richest}, labeling each instance as underspecified or not using a Qwen~\citep{bai2023qwen} model with Thinking Mode. \citet{huang2026richest} validate the quality of this annotation approach against human judgements.


The \textbf{NCQA} dataset \citep{nachshoni-etal-2025-consensus} comprises yes/no questions derived from fact-checking sources. While very small in size, this dataset offers a different source of aleatoric uncertainty, namely conflicting evidence rather than ambiguity in the questions.

\subsection{Models}

To generate the answers to the questions in the QA datasets, we employ five large language models representing three distinct paradigms: standard instruction tuning, diversity-oriented post-training, and ambiguity-aware fine-tuning.

For our standard baselines, we utilize \textbf{Llama-3.1-8B} \citep{grattafiori2024llama} and \textbf{Qwen3-14B} \citep{bai2023qwen}. We deploy Qwen3-14B in its non-thinking mode to avoid test-time compute scaling. 

To investigate interventions for aleatoric uncertainty, we evaluate \textbf{Spectrum-Llama-3.1-8B-v1} and \textbf{Spectrum-Qwen3-14B-v1} \citep{sorensen2025spectrum}. These variants maintain the base architectures but undergo Spectrum tuning to mitigate mode collapse. This enables them to preserve distributional coverage and generate diverse, valid responses to ambiguous prompts.

Finally, we benchmark against \textbf{A$^2$Search-7B-Instruct} \citep{zhang20252}, a model explicitly fine-tuned via reinforcement learning to recognize ambiguity, navigate conflicting evidence, and directly resolve underspecified queries. This model is based on Qwen2.5-7B \citep{bai2023qwen}.

\subsection{Output Evaluation}

To evaluate model performance, we employ an LLM-as-a-judge paradigm \citep{zheng2024judging}, using Gemma-2-2B-IT \citep{team2024gemma}. We choose Gemma to keep the judge architecturally independent from the Llama- and Qwen-based models being evaluated, avoiding same-family bias, while remaining small enough to score the full evaluation set efficiently. The judge is prompted to determine if the generated prediction is factually equivalent to the gold reference answers, allowing for minor surface-level variations. This verification is treated as a binary classification task, marking the prediction as correct if the normalized softmax probability of the generated ``yes'' token exceeds a predefined threshold ($\tau = 0.5$).

To evaluate factual consistency, we score responses using AlignScore \citep{zha2023alignscore}, which has been trained to evaluate factual correctness of a predicted answer against the gold. For instances containing multiple valid reference answers, we follow the common practice of evaluating the prediction against each reference independently, and assigning the maximum AlignScore achieved across all references for that instance~\citep{min-etal-2020-ambigqa, joshi-etal-2017-triviaqa}.

\subsection{Error Prediction Baseline} 
\label{sub:errpredbase}

The base model takes as input either a single scalar UQ metric or a combination of scalar UQ metrics $\mathbf{x}_f$, processed through a multi-layer perceptron (MLP) with layer normalisation and dropout to obtain a latent representation $\mathbf{z} = \text{MLP}(\mathbf{x}_f)$. This representation is then passed to the \textbf{Error Prediction} head, which estimates factual error likelihood via a non-linear classifier $\hat{c} = \text{head}_c(\mathbf{z})$. The model is optimized with a composite objective $\mathcal{L} = \mathcal{L}_{\text{BCE}} + \alpha \mathcal{L}_{\text{rank}}$, where $\mathcal{L}_{\text{BCE}}$ is the binary cross-entropy (BCE) loss and $\mathcal{L}_{\text{rank}} = \frac{1}{|P||N|}\sum_{i \in P, j \in N} \log(1 + \exp(\hat{c}_j - \hat{c}_i))$ is a pairwise softplus ranking loss over correct ($P$) and incorrect ($N$) samples with $\alpha = 0.05$, ensuring monotonic confidence ordering.

\subsection{Error Prediction Evaluation}

To evaluate the discriminative power of our uncertainty estimates, we utilize the \textbf{Area Under the Receiver Operating Characteristic curve (AUROC)} \citep{fawcett2006introduction}. AUROC provides a threshold-independent measure of a model's ability to rank incorrect generations as more uncertain than correct ones. 

Additionally, we report the \textbf{Prediction Rejection Ratio (PRR)} \citep{malinin2020uncertainty}, which quantifies the effectiveness of using uncertainty scores to abstain from answering. PRR normalizes the area under the accuracy-rejection curve by comparing the model's rejection strategy against both a random baseline and an optimal oracle, offering a standardized metric for selective generation quality.

Statistical significance of improvements over the baseline is assessed using a permutation test ($p < 0.05$).

\subsection{Implementation Details}

To quantify semantic uncertainty, we process each input in two generation phases. First, we establish a baseline response via greedy decoding, extracting both its sequence log-likelihood and the median-layer hidden state of its final generated token. Next, we draw $K=10$ stochastic samples $\{a_k\}_{k=1}^K$ using multinomial sampling ($T=1.0$). 

The experiments were carried out on NVIDIA GH200 Grace Hopper Superchips. Error Prediction models were trained for a maximum of 30 epochs, using early stopping when PRR does not improve, with patience of 5 epochs. 
We carried out a hyperparameter search with hidden dimensions in $\{128, 256, 512, 1024, 2048\}$ and learning rates in $\{3 \times 10^{-5}, 5 \times 10^{-5}, 1 \times 10^{-4}\}$.
We report the best score from the search for each model configuration. 
Following standard practice in the UQ literature \citep{vashurinCoCoA, farquhar2024detecting}, we report results on the validation splits of AmbigQA and TriviaQA, as the test set labels for both datasets are withheld. NCQA results are reported on its held-out test set.

\section{Results}

We present our main findings of the role of ambiguity in error prediction via UQ in Section~\ref{sub:resultserrpred}.
To lay the ground for them, in Section~\ref{sub:resultsqa} we demonstrate the model performance on the QA task, while Section~\ref{sub:resultsambig} presents the performance of ambiguity prediction models. 

\subsection{QA Performance}
\label{sub:resultsqa}

QA output evaluation reveals consistent differences between ambiguous and unambiguous instances across all models and datasets.
Table~\ref{tab:qascores} presents the LLM-as-a-judge and AlignScore values across models on the validation sets of all three datasets. 
All models score higher on TriviaQA and NCQA datasets as they are easier, given that TriviaQA is mostly unambiguous, and NCQA is a binary yes/no question set, while AmbigQA proves to be a much harder dataset to correctly classify.
Comparing ambiguous and unambiguous instances, we note that all models score much higher on both evaluation metrics for unambiguous instances. This result indicates that ambiguous ones are harder in general, because the evaluation metrics use a maximum score between all plausible reference answers, which gives an advantage to ambiguous questions. 
Notably, AlignScore values are lower across the board, indicating the stricter judgements of this metric compared to LLM-as-a-judge. For instance, while LLM-as-a-judge accepts the prediction \textit{`2001'} as a correct approximation of the reference \textit{`2001 fiscal year'}, AlignScore deems it not sufficiently precise. 

\begin{table}[t]
\centering
\scriptsize
\setlength{\tabcolsep}{2.5pt}
\begin{tabular}{llrrrrrr}
\toprule
& & \multicolumn{2}{c}{\textbf{All}} & \multicolumn{2}{c}{\textbf{Unamb.}} & \multicolumn{2}{c}{\textbf{Amb.}} \\
\cmidrule(lr){3-4} \cmidrule(lr){5-6} \cmidrule(lr){7-8}
\textbf{Model} & \textbf{Dataset} & \textbf{LLM} & \textbf{Align} & \textbf{LLM} & \textbf{Align} & \textbf{LLM} & \textbf{Align} \\
\midrule
\multirow{3}{*}{\shortstack[l]{Qwen3\\14B}}
  & AmbigQA       & 0.526 & 0.332 & 0.548 & 0.346 & 0.497 & 0.315 \\
  & TriviaQA      & 0.719 & 0.706 & 0.759 & 0.744 & 0.675 & 0.666 \\
  & NCQA & 0.920 & 0.920 & 0.556 & 0.556 & 1.000 & 1.000 \\
\midrule
\multirow{3}{*}{\shortstack[l]{Llama\\3.1-8B}}
  & AmbigQA       & 0.618 & 0.434 & 0.631 & 0.457 & 0.602 & 0.405 \\
  & TriviaQA      & 0.829 & 0.810 & 0.857 & 0.838 & 0.799 & 0.781 \\
  & NCQA & 0.850 & 0.850 & 0.167 & 0.167 & 1.000 & 1.000 \\
\midrule
\multirow{3}{*}{\shortstack[l]{Spectrum-\\Llama\\3.1-8B}}
  & AmbigQA       & 0.662 & 0.403 & 0.673 & 0.419 & 0.649 & 0.382 \\
  & TriviaQA      & 0.809 & 0.783 & 0.838 & 0.813 & 0.779 & 0.751 \\
  & NCQA & 0.860 & 0.860 & 0.222 & 0.222 & 1.000 & 1.000 \\
\midrule
\multirow{3}{*}{\shortstack[l]{Spectrum-\\Qwen3\\14B}}
  & AmbigQA       & 0.574 & 0.411 & 0.588 & 0.427 & 0.557 & 0.391 \\
  & TriviaQA      & 0.769 & 0.751 & 0.806 & 0.786 & 0.729 & 0.713 \\
  & NCQA & 0.910 & 0.910 & 0.500 & 0.500 & 1.000 & 1.000 \\
\midrule
\multirow{3}{*}{\shortstack[l]{A$^2$Search\\7B}}
  & AmbigQA       & 0.466 & 0.281 & 0.477 & 0.287 & 0.452 & 0.274 \\
  & TriviaQA      & 0.625 & 0.624 & 0.670 & 0.663 & 0.577 & 0.583 \\
  & NCQA & 0.910 & 0.910 & 0.500 & 0.500 & 1.000 & 1.000 \\
\bottomrule
\end{tabular}
\caption{LLM-judge (LLM) and AlignScore (Align) accuracy on validation splits by ambiguity (test for NCQA). }
\label{tab:qascores}
\end{table}

\subsection{Ambiguity Prediction}
\label{sub:resultsambig}

Table~\ref{tab:ambiguity_auroc} presents the ambiguity prediction scores for different models and datasets. Unsurprisingly, ambiguity prediction is most successful on AmbigQA, which has high quality human annotated ambiguity labels. In contrast, TriviaQA relies on LLM-generated annotations for underspecification (as discussed in Section~\ref{sub:data}). Furthermore, the limited scale of NCQA restricts the models' ability to extract robust patterns.

\begin{table}[t]
\centering
\scriptsize
\setlength{\tabcolsep}{4pt}
\begin{tabular}{lrrr}
\toprule
\textbf{Model} & \textbf{AmbigQA} & \textbf{TriviaQA} & \textbf{NCQA} \\
\midrule
  Llama-3.1-8B & 0.723 & 0.621 & 0.626 \\
  Qwen3-14B & 0.728 & 0.622 & 0.618 \\
  Spectrum-Llama-3.1-8B & 0.730 & 0.622 & 0.647 \\
  Spectrum-Qwen3-14B & 0.732 & 0.624 & 0.614 \\
  A$^2$Search-7B & 0.693 & 0.623 & 0.642 \\
\bottomrule
\end{tabular}
\caption{Ambiguity prediction AUROC per model and dataset (validation split, test for NCQA).}
\label{tab:ambiguity_auroc}
\end{table}

\subsection{Error Prediction}
\label{sub:resultserrpred}

The difference in UQ metric ability to predict model errors is presented in Table~\ref{tab:prr_qwen_alignscore}.\footnote{The results on all other models are in Appendix~\ref{app:error}.}
We find that ambiguity signal improves error prediction across models, datasets, UQ metrics, evaluation metrics, feature sets and architectures of error prediction models. 
Notably, the latent models often match or exceed their oracle counterparts despite using predicted rather than gold ambiguity labels. This suggests that the continuous ambiguity probabilities produced by the ambiguity classifier carry a more useful signal than the binary gold labels.

\begin{figure}
    \centering
    \includegraphics[width=\linewidth]{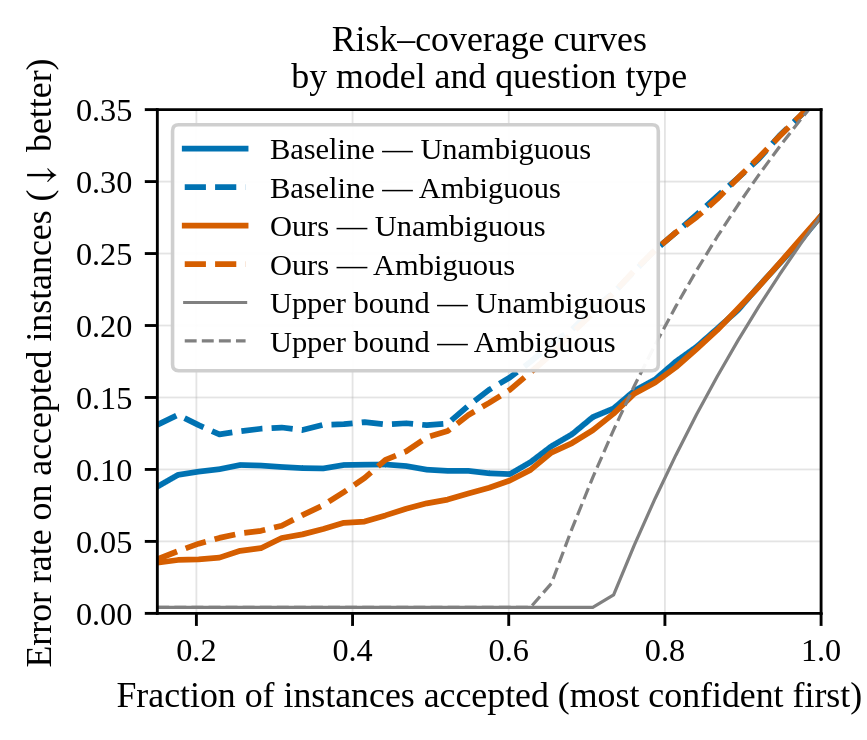}
    \caption{Risk–coverage curves for baseline and latent selective models on unambiguous and ambiguous questions, with upper bound (Qwen3-14B, TriviaQA).}
    \label{fig:fig2}
\end{figure}

Figure~\ref{fig:fig2} illustrates the improvements introduced by our proposed models. Each curve traces what happens as we accept progressively more of the model's answers, ordered from most to least confident. The y-axis shows the error rate among accepted answers, so a well-calibrated uncertainty measure should keep errors low until coverage is high. Curves are plotted separately for ambiguous and unambiguous questions, with the upper-bound lines showing what would be achievable if the uncertainty measure ranked every incorrect answer last.
First, the improvement appears across both ambiguous and unambiguous instances, indicating that the proposed architecture meaningfully encodes the two types of relationships, namely the one between epistemic uncertainty and error likelihood (unambiguous), and the one between a combination of aleatoric and epistemic uncertainty (ambiguous) and error likelihood. The full per-subset results across all models and datasets are provided in Appendix~\ref{app:error}, confirming this pattern holds beyond the model shown here.
Second, the improvement is most pronounced in the high-confidence area, where mistakes are the most detrimental.

Nonetheless, interesting differences emerge between the UQ metrics themselves. 
First, we find that Maximum Softmax Probability (MSP) is surprisingly robust. While complex semantic clustering methods are generally assumed to be superior, MSP frequently outperforms Semantic Entropy on inherently ambiguous datasets. This reinforces the premise that semantic clustering metrics degrade when valid aleatoric variations exist in the data.
Among the individual UQ features, CoCoA and SAR are in the lead. Notably, they still benefit from explicit ambiguity routing. 
Contrary to the design of MI as a UQ metric which is aware of aleatoric ambiguity, we find that the metric performs very poorly, especially on AmbigQA, corroborating the findings of previous research~\citep{tomov2025illusion}.
Crucially, we observe the highest overall error prediction performance when combining all UQ features into a single representation (\textit{all UQ}). Yet, even when the error prediction model has access to an ensemble of every UQ metric, applying our ambiguity signal via latent gating or selective prediction still yields consistent PRR improvements. This supports our core hypothesis that UQ metrics (even in combination) do not distinguish aleatoric and epistemic uncertainty, and our ambiguity classifier provides a complementary signal for establishing model reliability.

Comparing the different datasets, the NCQA improvements are the most dramatic due to the fact that, in contrast to other datasets, the ambiguous cases are the easy ones to solve given that both answers to a binary question would be deemed correct. 
We note that the improvements on NCQA should be interpreted with caution given the small training set of 234 instances. However, the results are reported on the held-out test split, and the pattern is consistent across feature sets and model families, suggesting it is not an artefact of overfitting to a single configuration.

Finally, models finetuned to represent multiple plausible answers (e.g. Spectrum-Qwen) are only marginally better at error prediction than their vanilla counterparts, suggesting such finetuning improves calibration alongside steerability but leaves room for further gains. These models still benefit from latent gating, indicating that ambiguity-aware finetuning and explicit ambiguity signal are not redundant.

\begin{table}[t]
\centering
\scriptsize
\setlength{\tabcolsep}{3pt}
\begin{tabular}{llrrrrr}
\toprule
\textbf{Dataset} & \textbf{Features} & \textbf{baseline} & \makecell{\textbf{latent}\\\textbf{gated}} & \makecell{\textbf{oracle}\\\textbf{gated}} & \makecell{\textbf{latent}\\\textbf{selective}} & \makecell{\textbf{oracle}\\\textbf{selective}} \\
\midrule
  \multirow{7}{*}{AmbigQA} & all UQ & 0.555 & \textbf{0.571} & \cellcolor{yellow!40}\textbf{0.575} & \textbf{0.562} & \textbf{0.537} \\
   & CoCoA & 0.538 & \textbf{0.553} & \textbf{0.553} & 0.524 & 0.526 \\
   & Sem. energy & 0.392 & \textbf{0.520} & \textbf{0.460} & \textbf{0.520} & \textbf{0.432} \\
   & Sem. entropy & 0.421 & \textbf{0.513} & \textbf{0.471} & \textbf{0.508} & \textbf{0.446} \\
   & SAR & 0.525 & \textbf{0.539} & \textbf{0.543} & 0.526 & 0.528 \\
   & MSP & 0.483 & \textbf{0.512} & \textbf{0.504} & \textbf{0.492} & 0.478 \\
   & MI & 0.014 & \textbf{0.078} & \textbf{0.058} & \textbf{0.078} & \textbf{0.058} \\
\midrule
  \multirow{7}{*}{TriviaQA} & all UQ & 0.787 & \textbf{0.795} & \textbf{0.788} & \cellcolor{yellow!40}\textbf{0.813} & \textbf{0.778} \\
   & CoCoA & 0.753 & \textbf{0.773} & 0.752 & \textbf{0.784} & \textbf{0.742} \\
   & Sem. energy & 0.754 & \textbf{0.778} & \textbf{0.757} & \textbf{0.793} & 0.748 \\
   & Sem. entropy & 0.641 & \textbf{0.759} & \textbf{0.673} & \textbf{0.761} & \textbf{0.671} \\
   & SAR & 0.735 & \textbf{0.773} & \textbf{0.739} & \textbf{0.786} & 0.729 \\
   & MSP & 0.738 & \textbf{0.758} & \textbf{0.738} & \textbf{0.771} & \textbf{0.725} \\
   & MI & 0.204 & \textbf{0.568} & 0.196 & \textbf{0.569} & 0.197 \\
\midrule
  \multirow{7}{*}{NCQA} & all UQ & 0.499 & \textbf{0.577} & \textbf{0.967} & \textbf{0.534} & \textbf{0.973} \\
   & CoCoA & 0.476 & \textbf{0.606} & \textbf{0.963} & \textbf{0.520} & \textbf{0.964} \\
   & Sem. energy & 0.468 & \textbf{0.489} & \textbf{0.971} & 0.436 & \textbf{0.972} \\
   & Sem. entropy & 0.086 & \textbf{0.533} & \textbf{0.968} & -0.545 & \textbf{0.956} \\
   & SAR & 0.684 & 0.644 & \textbf{0.984} & -0.319 & \cellcolor{yellow!40}\textbf{0.986} \\
   & MSP & 0.444 & \textbf{0.623} & \textbf{0.951} & \textbf{0.645} & \textbf{0.971} \\
   & MI & -0.160 & \textbf{0.576} & \textbf{0.946} & \textbf{0.576} & \textbf{0.945} \\
\bottomrule
\end{tabular}
\caption{PRR on validation split (test for NCQA) for Qwen3-14B (AlignScore supervision). \textbf{Bold} = significant improvement over baseline ($p < 0.05$, test set). \colorbox{yellow!40}{Highlight} = best PRR for model/dataset.}
\label{tab:prr_qwen_alignscore}
\end{table}

To illustrate, consider the pair of questions in Table~\ref{tab:ambiguity_example}.
The top question is
underspecified, where valid answers include 256 (total levels) and 21 (levels before difficulty plateaus).
The question about sonnets has one correct answer (14 lines).
A UQ-based model trained on semantic entropy treats output diversity as a
signal of likely error, assigning 73.8\% predicted error to the ambiguous
question.  Our latent selective model disentangles underspecification from
error, assigning near-zero predicted error (0.0\%) while giving a similarly
low prediction for the unambiguous sonnet (1.2\%).  The ambiguity model makes correct predictions (0.527 for Pac-Man versus 0.440 for the sonnet), though the small margin reflects the difficulty of predicting ambiguity discussed in Section~\ref{sub:resultsambig}.

\begin{table}[ht]
\centering
\scriptsize
\renewcommand{\arraystretch}{1.2}
\setlength{\tabcolsep}{3pt}
\begin{tabular}{l l r r r r r r r}
\toprule
\textbf{Amb.} & \textbf{Question} & \textbf{Acc} & \textbf{Base} & \textbf{Ours}& $\hat{p}_\text{amb}$ \\
\midrule
\tick & How many levels are there in Pac-Man? & 1.00 & 0.74 & 0.00 & 0.53 \\
\cross & How many lines are there in a sonnet? & 1.00 & 0.03 & 0.01 & 0.44 \\
\bottomrule
\end{tabular}
\caption{Predicted error probability. Acc\ = alignscore of the
  most likely response. Base\ = semantic entropy only. Ours = latent selective.
  $\hat{p}_\text{amb}$\ = predicted ambiguity.}
\label{tab:ambiguity_example}
\end{table}

\section{Conclusion}

In this paper we claim that ambiguity signal is useful for predicting errors from uncertainty, due to the fact that uncertainty encompasses both epistemic and aleatoric sources. 
Experimental results show that ambiguity information is beneficial regardless of the model, UQ metric, dataset, feature set and architecture choice for error prediction. 
We show that using UQ metrics along with ambiguity can yield high error prediction performance, up to 0.879 PRR and 0.904 AUROC on TriviaQA. 
These results contribute to improving the reliability of LLMs. 
Future work could extend this framework to retrieval-augmented settings, where retrieved context introduces a further source of aleatoric uncertainty, or to settings where ambiguity labels must be induced fully unsupervised.

\section*{Limitations}

The study only covers textual data, and all questions and answers are in the English language, hence further work would need to establish whether the discovered trends generalise to other languages, multilingual models and multimodal models. 

Moreover, our latent models require ambiguity-labelled data at training time, which may not be available for all domains and tasks. Extending our approach to settings where ambiguity labels must be fully automatically induced remains an important direction for future work.

Furthermore, the study focuses on semantic uncertainty quantification metrics, which is one of many types of measures of uncertainty. While the semantically rich metrics achieve state-of-the-art results and are therefore the most interesting to study, other work could explore whether other UQ metrics would have a similar relationship to ambiguity. 

In addition, hyperparameter selection is performed by evaluating PRR on the same validation set used for final reporting. While this is a common practice in error prediction research due to the absence of held-out test splits for AmbigQA and TriviaQA (whose test labels are withheld for leaderboard evaluation), it may lead to optimistic reported scores; results should be interpreted accordingly.

Finally, this research paper is limited to closed-book QA, which restricts the knowledge available to the model to that which was present in the training data. It would be of great interest to extend the work to a retrieval-augmented-generation setup and compare the uncertainty and ambiguity relationship to a setup where more information is available via the retrieval process, especially the cases where the retrieved data contradicts internal knowledge.

\section*{Acknowledgments}

\bibliography{custom}

\appendix

\section{Full Feature Correlation with Ambiguity Labels}
\label{app:correlation}

Table~\ref{tab:feat_corr_other} extends the per-feature AUROC analysis of Table~\ref{tab:feat_corr_llama} to the remaining four models. The pattern observed for Llama-3.1-8B holds across model families: representation features (last-layer embeddings and mid-layer residuals) provide the strongest standalone signal for ambiguity on AmbigQA, while scalar UQ and semantic-cluster features cluster near chance, with entropy- and diversity-based features slightly outperforming concentration-based ones.

\begin{table*}[t]
\centering
\tiny
\setlength{\tabcolsep}{2pt}
\begin{tabular}{llrrrrrrrrrrrrrr}
\toprule
& & \multicolumn{2}{c}{\textbf{Internal Repr.}} & \multicolumn{5}{c}{\textbf{UQ}} & \multicolumn{7}{c}{\textbf{Sem. Clust.}} \\
\cmidrule(lr){3-4} \cmidrule(lr){5-9} \cmidrule(lr){10-16}
\textbf{Model} & \textbf{Dataset} & \textbf{Emb} & \textbf{Res} & \textbf{SE} & \textbf{SEng} & \textbf{MSP} & \textbf{SAR} & \textbf{MI} & \textbf{CoCoA} & \textbf{\#Cl} & \textbf{ClEnt} & \textbf{MxCP} & \textbf{ENA} & \textbf{PG2} & \textbf{Top2} \\
\midrule
  \multirow{3}{*}{\shortstack[l]{Spectrum-Llama\\3.1-8B}} & AmbigQA & 0.726 & 0.741 & 0.541 & 0.538 & 0.529 & 0.510 & 0.542 & 0.532 & 0.536 & 0.541 & 0.459 & 0.541 & 0.461 & 0.460 \\
   & NCQA & 0.501 & 0.603 & 0.507 & 0.490 & 0.572 & 0.519 & 0.541 & 0.508 & 0.556 & 0.507 & 0.491 & 0.507 & 0.494 & 0.465 \\
   & TriviaQA & 0.576 & 0.570 & 0.560 & 0.566 & 0.556 & 0.551 & 0.565 & 0.563 & 0.565 & 0.561 & 0.443 & 0.561 & 0.448 & 0.437 \\
\midrule
  \multirow{3}{*}{\shortstack[l]{Qwen3\\14B}} & AmbigQA & 0.718 & 0.715 & 0.525 & 0.507 & 0.484 & 0.477 & 0.458 & 0.522 & 0.518 & 0.527 & 0.469 & 0.527 & 0.468 & 0.480 \\
   & NCQA & 0.432 & 0.442 & 0.450 & 0.495 & 0.503 & 0.464 & 0.514 & 0.499 & 0.481 & 0.478 & 0.522 & 0.478 & 0.522 & 0.500 \\
   & TriviaQA & 0.588 & 0.594 & 0.546 & 0.555 & 0.554 & 0.555 & 0.463 & 0.556 & 0.549 & 0.548 & 0.453 & 0.548 & 0.455 & 0.454 \\
\midrule
  \multirow{3}{*}{\shortstack[l]{Spectrum-Qwen3\\14B}} & AmbigQA & 0.730 & 0.733 & 0.538 & 0.524 & 0.519 & 0.501 & 0.541 & 0.526 & 0.535 & 0.538 & 0.463 & 0.538 & 0.464 & 0.462 \\
   & NCQA & 0.305 & 0.401 & 0.463 & 0.457 & 0.528 & 0.474 & 0.505 & 0.513 & 0.568 & 0.463 & 0.566 & 0.463 & 0.577 & 0.402 \\
   & TriviaQA & 0.571 & 0.579 & 0.565 & 0.567 & 0.562 & 0.539 & 0.567 & 0.570 & 0.566 & 0.565 & 0.439 & 0.565 & 0.443 & 0.436 \\
\midrule
  \multirow{3}{*}{\shortstack[l]{A$^2$Search\\7B}} & AmbigQA & 0.707 & 0.714 & 0.494 & 0.476 & 0.451 & 0.458 & 0.460 & 0.483 & 0.485 & 0.494 & 0.502 & 0.494 & 0.500 & 0.507 \\
   & NCQA & 0.634 & 0.514 & 0.515 & 0.546 & 0.608 & 0.552 & 0.491 & 0.598 & 0.485 & 0.486 & 0.514 & 0.486 & 0.514 & 0.513 \\
   & TriviaQA & 0.588 & 0.584 & 0.540 & 0.545 & 0.542 & 0.543 & 0.468 & 0.543 & 0.540 & 0.539 & 0.462 & 0.539 & 0.464 & 0.468 \\
\bottomrule
\end{tabular}
\caption{Same as Table~\ref{tab:feat_corr_llama} for Spectrum-Llama, Qwen3-14B, Spectrum-Qwen3-14B and A$^2$Search-7B. 
AUROC of individual features for predicting ambiguity across all models and datasets (validation split). Scalar features evaluated directly; representation features use 5-fold CV logistic regression.}
\label{tab:feat_corr_other}
\end{table*}

\pagebreak

\section{Full Error Prediction Results}
\label{app:error}

Tables~\ref{tab:prr_ambigqa_alignscore}--\ref{tab:auroc_conflictingqa_llm} report PRR and AUROC for error prediction across all five models, three datasets, and two correctness supervision signals (AlignScore and LLM-judge), with per-subset scores for ambiguous and unambiguous instances shown beneath each main value. Across this full grid, the pattern reported in Section~\ref{sub:resultserrpred} for Qwen3-14B holds: ambiguity-aware models (gated or selective, latent or oracle) improve over the UQ-only baseline for the large majority of model--feature combinations, with the latent variants typically tracking or exceeding the oracle. The few non-significant or negative entries are concentrated on NCQA, where the small training set yields high variance, and on MI, whose baseline performance is already close to chance.

\begin{table*}[t]
\centering
\footnotesize
\begin{adjustbox}{max width=\textwidth, max totalheight=0.92\textheight, keepaspectratio}
\setlength{\tabcolsep}{4pt}
\begin{tabular}{llrrrrr}
\toprule
\textbf{Model} & \textbf{Features} & \textbf{baseline} & \makecell{\textbf{latent}\\\textbf{gated}} & \makecell{\textbf{oracle}\\\textbf{gated}} & \makecell{\textbf{latent}\\\textbf{selective}} & \makecell{\textbf{oracle}\\\textbf{selective}} \\
\midrule
  \multirow{7}{*}{Llama-3.1-8B} & all UQ & \makecell[r]{0.632\\{\scriptsize 0.722}\\{\scriptsize 0.504}} & \makecell[r]{0.633\\{\scriptsize 0.722}\\{\scriptsize 0.502}} & \cellcolor{yellow!40}\makecell[r]{\textbf{0.651}\\{\scriptsize 0.727}\\{\scriptsize 0.511}} & \makecell[r]{0.620\\{\scriptsize 0.708}\\{\scriptsize 0.483}} & \makecell[r]{0.627\\{\scriptsize 0.721}\\{\scriptsize 0.502}} \\
   & CoCoA & \makecell[r]{0.615\\{\scriptsize 0.706}\\{\scriptsize 0.490}} & \makecell[r]{0.613\\{\scriptsize 0.703}\\{\scriptsize 0.487}} & \makecell[r]{\textbf{0.629}\\{\scriptsize 0.706}\\{\scriptsize 0.490}} & \makecell[r]{0.611\\{\scriptsize 0.700}\\{\scriptsize 0.481}} & \makecell[r]{0.614\\{\scriptsize 0.706}\\{\scriptsize 0.490}} \\
   & Semantic energy & \makecell[r]{0.593\\{\scriptsize 0.684}\\{\scriptsize 0.462}} & \makecell[r]{0.596\\{\scriptsize 0.682}\\{\scriptsize 0.463}} & \makecell[r]{\textbf{0.606}\\{\scriptsize 0.682}\\{\scriptsize 0.461}} & \makecell[r]{0.585\\{\scriptsize 0.668}\\{\scriptsize 0.450}} & \makecell[r]{0.585\\{\scriptsize 0.682}\\{\scriptsize 0.461}} \\
   & Semantic entropy & \makecell[r]{0.595\\{\scriptsize 0.677}\\{\scriptsize 0.463}} & \makecell[r]{0.595\\{\scriptsize 0.680}\\{\scriptsize 0.463}} & \makecell[r]{\textbf{0.611}\\{\scriptsize 0.680}\\{\scriptsize 0.463}} & \makecell[r]{0.583\\{\scriptsize 0.665}\\{\scriptsize 0.446}} & \makecell[r]{0.589\\{\scriptsize 0.680}\\{\scriptsize 0.462}} \\
\bottomrule
\end{tabular}
\end{adjustbox}
\caption{PRR on AmbigQA (validation split), AlignScore supervision. \textbf{Bold} = significant improvement over baseline ($p < 0.05$, test set). \colorbox{yellow!40}{Highlight} = best PRR for model. }
\label{tab:prr_ambigqa_alignscore}
\end{table*}

\begin{table*}[t]
\centering
\footnotesize
\begin{adjustbox}{max width=\textwidth, max totalheight=0.92\textheight, keepaspectratio}
\setlength{\tabcolsep}{4pt}
\begin{tabular}{llrrrrr}
\toprule
\textbf{Model} & \textbf{Features} & \textbf{baseline} & \makecell{\textbf{latent}\\\textbf{gated}} & \makecell{\textbf{oracle}\\\textbf{gated}} & \makecell{\textbf{latent}\\\textbf{selective}} & \makecell{\textbf{oracle}\\\textbf{selective}} \\
\midrule
  \multirow{7}{*}{Llama-3.1-8B} & all UQ & \makecell[r]{0.848\\{\scriptsize 0.858}\\{\scriptsize 0.833}} & \makecell[r]{0.849\\{\scriptsize 0.859}\\{\scriptsize 0.833}} & \makecell[r]{0.848\\{\scriptsize 0.856}\\{\scriptsize 0.834}} & \cellcolor{yellow!40}\makecell[r]{0.850\\{\scriptsize 0.859}\\{\scriptsize 0.835}} & \makecell[r]{0.839\\{\scriptsize 0.857}\\{\scriptsize 0.832}} \\
   & CoCoA & \makecell[r]{0.829\\{\scriptsize 0.838}\\{\scriptsize 0.814}} & \makecell[r]{0.831\\{\scriptsize 0.838}\\{\scriptsize 0.816}} & \makecell[r]{0.830\\{\scriptsize 0.838}\\{\scriptsize 0.814}} & \makecell[r]{0.831\\{\scriptsize 0.838}\\{\scriptsize 0.817}} & \makecell[r]{0.817\\{\scriptsize 0.838}\\{\scriptsize 0.814}} \\
   & Semantic energy & \makecell[r]{0.814\\{\scriptsize 0.822}\\{\scriptsize 0.799}} & \makecell[r]{0.818\\{\scriptsize 0.824}\\{\scriptsize 0.804}} & \makecell[r]{0.815\\{\scriptsize 0.822}\\{\scriptsize 0.799}} & \makecell[r]{0.819\\{\scriptsize 0.825}\\{\scriptsize 0.805}} & \makecell[r]{0.802\\{\scriptsize 0.822}\\{\scriptsize 0.799}} \\
   & Semantic entropy & \makecell[r]{0.813\\{\scriptsize 0.815}\\{\scriptsize 0.801}} & \makecell[r]{\textbf{0.820}\\{\scriptsize 0.828}\\{\scriptsize 0.806}} & \makecell[r]{0.815\\{\scriptsize 0.815}\\{\scriptsize 0.800}} & \makecell[r]{\textbf{0.822}\\{\scriptsize 0.829}\\{\scriptsize 0.808}} & \makecell[r]{0.804\\{\scriptsize 0.815}\\{\scriptsize 0.800}} \\
\bottomrule
\end{tabular}
\end{adjustbox}
\caption{PRR on TriviaQA (validation split), AlignScore supervision. \textbf{Bold} = significant improvement over baseline ($p < 0.05$, test set). \colorbox{yellow!40}{Highlight} = best PRR for model.}
\label{tab:prr_triviaqa_alignscore}
\end{table*}

\begin{table*}[t]
\centering
\footnotesize
\begin{adjustbox}{max width=\textwidth, max totalheight=0.92\textheight, keepaspectratio}
\setlength{\tabcolsep}{4pt}
\begin{tabular}{llrrrrr}
\toprule
\textbf{Model} & \textbf{Features} & \textbf{baseline} & \makecell{\textbf{latent}\\\textbf{gated}} & \makecell{\textbf{oracle}\\\textbf{gated}} & \makecell{\textbf{latent}\\\textbf{selective}} & \makecell{\textbf{oracle}\\\textbf{selective}} \\
\midrule
  \multirow{7}{*}{Llama-3.1-8B} & all UQ & \makecell[r]{0.330\\{\scriptsize 0.052}\\{\scriptsize 0.000}} & \makecell[r]{0.369\\{\scriptsize -0.052}\\{\scriptsize 0.000}} & \makecell[r]{\textbf{0.991}\\{\scriptsize 0.416}\\{\scriptsize 0.000}} & \makecell[r]{0.442\\{\scriptsize 0.387}\\{\scriptsize 0.000}} & \makecell[r]{\textbf{0.991}\\{\scriptsize 0.485}\\{\scriptsize 0.000}} \\
   & CoCoA & \makecell[r]{0.334\\{\scriptsize 0.299}\\{\scriptsize 0.000}} & \makecell[r]{0.311\\{\scriptsize 0.321}\\{\scriptsize 0.000}} & \makecell[r]{\textbf{0.987}\\{\scriptsize 0.209}\\{\scriptsize 0.000}} & \makecell[r]{0.267\\{\scriptsize 0.249}\\{\scriptsize 0.000}} & \makecell[r]{\textbf{0.987}\\{\scriptsize 0.315}\\{\scriptsize 0.000}} \\
   & Semantic energy & \makecell[r]{0.306\\{\scriptsize 0.296}\\{\scriptsize 0.000}} & \makecell[r]{0.289\\{\scriptsize 0.209}\\{\scriptsize 0.000}} & \makecell[r]{\textbf{0.992}\\{\scriptsize 0.558}\\{\scriptsize 0.000}} & \makecell[r]{0.287\\{\scriptsize 0.284}\\{\scriptsize 0.000}} & \makecell[r]{\textbf{0.988}\\{\scriptsize 0.479}\\{\scriptsize 0.000}} \\
   & Semantic entropy & \makecell[r]{0.378\\{\scriptsize -0.025}\\{\scriptsize 0.000}} & \makecell[r]{0.293\\{\scriptsize 0.284}\\{\scriptsize 0.000}} & \makecell[r]{\textbf{0.988}\\{\scriptsize 0.350}\\{\scriptsize 0.000}} & \makecell[r]{0.266\\{\scriptsize 0.249}\\{\scriptsize 0.000}} & \makecell[r]{\textbf{0.989}\\{\scriptsize 0.357}\\{\scriptsize 0.000}} \\
\bottomrule
\end{tabular}
\end{adjustbox}
\caption{PRR on NCQA (test split), AlignScore supervision. \textbf{Bold} = significant improvement over baseline ($p < 0.05$, test set). \colorbox{yellow!40}{Highlight} = best PRR for model.}
\label{tab:prr_conflictingqa_alignscore}
\end{table*}

\begin{table*}[t]
\centering
\footnotesize
\begin{adjustbox}{max width=\textwidth, max totalheight=0.92\textheight, keepaspectratio}
\setlength{\tabcolsep}{4pt}
\begin{tabular}{llrrrrr}
\toprule
\textbf{Model} & \textbf{Features} & \textbf{baseline} & \makecell{\textbf{latent}\\\textbf{gated}} & \makecell{\textbf{oracle}\\\textbf{gated}} & \makecell{\textbf{latent}\\\textbf{selective}} & \makecell{\textbf{oracle}\\\textbf{selective}} \\
\midrule
  \multirow{7}{*}{Llama-3.1-8B} & all UQ & \makecell[r]{0.806\\{\scriptsize 0.846}\\{\scriptsize 0.754}} & \makecell[r]{0.806\\{\scriptsize 0.845}\\{\scriptsize 0.753}} & \cellcolor{yellow!40}\makecell[r]{0.808\\{\scriptsize 0.845}\\{\scriptsize 0.755}} & \makecell[r]{0.805\\{\scriptsize 0.838}\\{\scriptsize 0.745}} & \makecell[r]{0.806\\{\scriptsize 0.846}\\{\scriptsize 0.753}} \\
   & CoCoA & \makecell[r]{0.792\\{\scriptsize 0.835}\\{\scriptsize 0.737}} & \makecell[r]{0.791\\{\scriptsize 0.834}\\{\scriptsize 0.738}} & \makecell[r]{0.795\\{\scriptsize 0.836}\\{\scriptsize 0.737}} & \makecell[r]{0.778\\{\scriptsize 0.825}\\{\scriptsize 0.726}} & \makecell[r]{0.792\\{\scriptsize 0.835}\\{\scriptsize 0.737}} \\
   & Semantic energy & \makecell[r]{0.777\\{\scriptsize 0.817}\\{\scriptsize 0.727}} & \makecell[r]{\textbf{0.779}\\{\scriptsize 0.817}\\{\scriptsize 0.727}} & \makecell[r]{\textbf{0.779}\\{\scriptsize 0.816}\\{\scriptsize 0.726}} & \makecell[r]{0.777\\{\scriptsize 0.817}\\{\scriptsize 0.727}} & \makecell[r]{0.776\\{\scriptsize 0.816}\\{\scriptsize 0.726}} \\
   & Semantic entropy & \makecell[r]{0.780\\{\scriptsize 0.818}\\{\scriptsize 0.732}} & \makecell[r]{0.781\\{\scriptsize 0.819}\\{\scriptsize 0.731}} & \makecell[r]{\textbf{0.783}\\{\scriptsize 0.818}\\{\scriptsize 0.732}} & \makecell[r]{0.781\\{\scriptsize 0.819}\\{\scriptsize 0.730}} & \makecell[r]{0.780\\{\scriptsize 0.818}\\{\scriptsize 0.732}} \\
\bottomrule
\end{tabular}
\end{adjustbox}
\caption{AUROC (error prediction) on AmbigQA (validation split), AlignScore supervision. \textbf{Bold} = significant improvement over baseline ($p < 0.05$, test set). \colorbox{yellow!40}{Highlight} = best AUROC for model. }
\label{tab:auroc_ambigqa_alignscore}
\end{table*}

\begin{table*}[t]
\centering
\footnotesize
\begin{adjustbox}{max width=\textwidth, max totalheight=0.92\textheight, keepaspectratio}
\setlength{\tabcolsep}{4pt}
\begin{tabular}{llrrrrr}
\toprule
\textbf{Model} & \textbf{Features} & \textbf{baseline} & \makecell{\textbf{latent}\\\textbf{gated}} & \makecell{\textbf{oracle}\\\textbf{gated}} & \makecell{\textbf{latent}\\\textbf{selective}} & \makecell{\textbf{oracle}\\\textbf{selective}} \\
\midrule
  \multirow{7}{*}{Llama-3.1-8B} & all UQ & \makecell[r]{0.886\\{\scriptsize 0.894}\\{\scriptsize 0.876}} & \makecell[r]{\textbf{0.887}\\{\scriptsize 0.894}\\{\scriptsize 0.876}} & \makecell[r]{0.886\\{\scriptsize 0.893}\\{\scriptsize 0.876}} & \cellcolor{yellow!40}\makecell[r]{\textbf{0.888}\\{\scriptsize 0.895}\\{\scriptsize 0.877}} & \makecell[r]{0.886\\{\scriptsize 0.893}\\{\scriptsize 0.876}} \\
   & CoCoA & \makecell[r]{0.874\\{\scriptsize 0.881}\\{\scriptsize 0.862}} & \makecell[r]{0.874\\{\scriptsize 0.881}\\{\scriptsize 0.864}} & \makecell[r]{0.874\\{\scriptsize 0.881}\\{\scriptsize 0.862}} & \makecell[r]{0.875\\{\scriptsize 0.881}\\{\scriptsize 0.863}} & \makecell[r]{0.874\\{\scriptsize 0.881}\\{\scriptsize 0.862}} \\
   & Semantic energy & \makecell[r]{0.865\\{\scriptsize 0.872}\\{\scriptsize 0.853}} & \makecell[r]{\textbf{0.866}\\{\scriptsize 0.873}\\{\scriptsize 0.855}} & \makecell[r]{0.865\\{\scriptsize 0.872}\\{\scriptsize 0.853}} & \makecell[r]{\textbf{0.867}\\{\scriptsize 0.874}\\{\scriptsize 0.856}} & \makecell[r]{0.865\\{\scriptsize 0.872}\\{\scriptsize 0.853}} \\
   & Semantic entropy & \makecell[r]{0.868\\{\scriptsize 0.874}\\{\scriptsize 0.858}} & \makecell[r]{\textbf{0.870}\\{\scriptsize 0.876}\\{\scriptsize 0.860}} & \makecell[r]{0.868\\{\scriptsize 0.874}\\{\scriptsize 0.858}} & \makecell[r]{\textbf{0.871}\\{\scriptsize 0.876}\\{\scriptsize 0.861}} & \makecell[r]{0.868\\{\scriptsize 0.874}\\{\scriptsize 0.858}} \\
\bottomrule
\end{tabular}
\end{adjustbox}
\caption{AUROC (error prediction) on TriviaQA (validation split), AlignScore supervision. \textbf{Bold} = significant improvement over baseline ($p < 0.05$, test set). \colorbox{yellow!40}{Highlight} = best AUROC for model. }
\label{tab:auroc_triviaqa_alignscore}
\end{table*}

\begin{table*}[t]
\centering
\footnotesize
\begin{adjustbox}{max width=\textwidth, max totalheight=0.92\textheight, keepaspectratio}
\setlength{\tabcolsep}{4pt}
\begin{tabular}{llrrrrr}
\toprule
\textbf{Model} & \textbf{Features} & \textbf{baseline} & \makecell{\textbf{latent}\\\textbf{gated}} & \makecell{\textbf{oracle}\\\textbf{gated}} & \makecell{\textbf{latent}\\\textbf{selective}} & \makecell{\textbf{oracle}\\\textbf{selective}} \\
\midrule
  \multirow{7}{*}{Llama-3.1-8B} & all UQ & \makecell[r]{0.615\\{\scriptsize 0.533}\\{\scriptsize 0.500}} & \makecell[r]{0.599\\{\scriptsize 0.578}\\{\scriptsize 0.500}} & \makecell[r]{\textbf{0.991}\\{\scriptsize 0.756}\\{\scriptsize 0.500}} & \makecell[r]{0.649\\{\scriptsize 0.756}\\{\scriptsize 0.500}} & \makecell[r]{0.578\\{\scriptsize 0.756}\\{\scriptsize 0.500}} \\
   & CoCoA & \makecell[r]{0.598\\{\scriptsize 0.622}\\{\scriptsize 0.500}} & \makecell[r]{0.584\\{\scriptsize 0.622}\\{\scriptsize 0.500}} & \makecell[r]{\textbf{0.988}\\{\scriptsize 0.644}\\{\scriptsize 0.500}} & \makecell[r]{0.542\\{\scriptsize 0.600}\\{\scriptsize 0.500}} & \makecell[r]{0.610\\{\scriptsize 0.644}\\{\scriptsize 0.500}} \\
   & Semantic energy & \makecell[r]{0.569\\{\scriptsize 0.622}\\{\scriptsize 0.500}} & \makecell[r]{0.591\\{\scriptsize 0.644}\\{\scriptsize 0.500}} & \cellcolor{yellow!40}\makecell[r]{\textbf{0.992}\\{\scriptsize 0.778}\\{\scriptsize 0.500}} & \makecell[r]{0.569\\{\scriptsize 0.578}\\{\scriptsize 0.500}} & \makecell[r]{0.508\\{\scriptsize 0.667}\\{\scriptsize 0.500}} \\
   & Semantic entropy & \makecell[r]{0.620\\{\scriptsize 0.556}\\{\scriptsize 0.500}} & \makecell[r]{0.581\\{\scriptsize 0.622}\\{\scriptsize 0.500}} & \makecell[r]{\textbf{0.989}\\{\scriptsize 0.689}\\{\scriptsize 0.500}} & \makecell[r]{0.547\\{\scriptsize 0.622}\\{\scriptsize 0.500}} & \makecell[r]{0.490\\{\scriptsize 0.711}\\{\scriptsize 0.500}} \\
\bottomrule
\end{tabular}
\end{adjustbox}
\caption{AUROC (error prediction) on NCQA (test split), AlignScore supervision. \textbf{Bold} = significant improvement over baseline ($p < 0.05$, test set). \colorbox{yellow!40}{Highlight} = best AUROC for model. }
\label{tab:auroc_conflictingqa_alignscore}
\end{table*}

\begin{table*}[t]
\centering
\footnotesize
\begin{adjustbox}{max width=\textwidth, max totalheight=0.92\textheight, keepaspectratio}
\setlength{\tabcolsep}{4pt}
\begin{tabular}{llrrrrr}
\toprule
\textbf{Model} & \textbf{Features} & \textbf{baseline} & \makecell{\textbf{latent}\\\textbf{gated}} & \makecell{\textbf{oracle}\\\textbf{gated}} & \makecell{\textbf{latent}\\\textbf{selective}} & \makecell{\textbf{oracle}\\\textbf{selective}} \\
\midrule
  \multirow{7}{*}{Llama-3.1-8B} & all UQ & \makecell[r]{0.696\\{\scriptsize 0.731}\\{\scriptsize 0.652}} & \makecell[r]{0.695\\{\scriptsize 0.728}\\{\scriptsize 0.651}} & \cellcolor{yellow!40}\makecell[r]{0.701\\{\scriptsize 0.734}\\{\scriptsize 0.650}} & \makecell[r]{0.681\\{\scriptsize 0.709}\\{\scriptsize 0.640}} & \makecell[r]{0.681\\{\scriptsize 0.727}\\{\scriptsize 0.651}} \\
   & CoCoA & \makecell[r]{0.644\\{\scriptsize 0.683}\\{\scriptsize 0.592}} & \makecell[r]{\textbf{0.651}\\{\scriptsize 0.688}\\{\scriptsize 0.601}} & \makecell[r]{0.649\\{\scriptsize 0.684}\\{\scriptsize 0.592}} & \makecell[r]{0.646\\{\scriptsize 0.680}\\{\scriptsize 0.596}} & \makecell[r]{0.626\\{\scriptsize 0.683}\\{\scriptsize 0.592}} \\
   & Semantic energy & \makecell[r]{0.618\\{\scriptsize 0.652}\\{\scriptsize 0.571}} & \makecell[r]{0.620\\{\scriptsize 0.654}\\{\scriptsize 0.573}} & \makecell[r]{0.620\\{\scriptsize 0.651}\\{\scriptsize 0.571}} & \makecell[r]{0.611\\{\scriptsize 0.640}\\{\scriptsize 0.567}} & \makecell[r]{0.600\\{\scriptsize 0.651}\\{\scriptsize 0.571}} \\
   & Semantic entropy & \makecell[r]{0.614\\{\scriptsize 0.651}\\{\scriptsize 0.557}} & \makecell[r]{\textbf{0.618}\\{\scriptsize 0.654}\\{\scriptsize 0.567}} & \makecell[r]{0.616\\{\scriptsize 0.651}\\{\scriptsize 0.557}} & \makecell[r]{0.604\\{\scriptsize 0.639}\\{\scriptsize 0.553}} & \makecell[r]{0.592\\{\scriptsize 0.651}\\{\scriptsize 0.557}} \\
\bottomrule
\end{tabular}
\end{adjustbox}
\caption{PRR on AmbigQA (validation split), LLM-judge supervision. \textbf{Bold} = significant improvement over baseline ($p < 0.05$, test set). \colorbox{yellow!40}{Highlight} = best PRR for model. }
\label{tab:prr_ambigqa_llm}
\end{table*}

\begin{table*}[t]
\centering
\footnotesize
\begin{adjustbox}{max width=\textwidth, max totalheight=0.92\textheight, keepaspectratio}
\setlength{\tabcolsep}{4pt}
\begin{tabular}{llrrrrr}
\toprule
\textbf{Model} & \textbf{Features} & \textbf{baseline} & \makecell{\textbf{latent}\\\textbf{gated}} & \makecell{\textbf{oracle}\\\textbf{gated}} & \makecell{\textbf{latent}\\\textbf{selective}} & \makecell{\textbf{oracle}\\\textbf{selective}} \\
\midrule
  \multirow{7}{*}{Llama-3.1-8B} & all UQ & \makecell[r]{0.878\\{\scriptsize 0.891}\\{\scriptsize 0.861}} & \makecell[r]{0.878\\{\scriptsize 0.892}\\{\scriptsize 0.860}} & \makecell[r]{0.878\\{\scriptsize 0.890}\\{\scriptsize 0.860}} & \cellcolor{yellow!40}\makecell[r]{0.879\\{\scriptsize 0.892}\\{\scriptsize 0.860}} & \makecell[r]{0.869\\{\scriptsize 0.891}\\{\scriptsize 0.860}} \\
   & CoCoA & \makecell[r]{0.850\\{\scriptsize 0.860}\\{\scriptsize 0.834}} & \makecell[r]{\textbf{0.851}\\{\scriptsize 0.861}\\{\scriptsize 0.835}} & \makecell[r]{0.850\\{\scriptsize 0.860}\\{\scriptsize 0.833}} & \makecell[r]{\textbf{0.851}\\{\scriptsize 0.862}\\{\scriptsize 0.834}} & \makecell[r]{0.842\\{\scriptsize 0.860}\\{\scriptsize 0.833}} \\
   & Semantic energy & \makecell[r]{0.835\\{\scriptsize 0.848}\\{\scriptsize 0.815}} & \makecell[r]{\textbf{0.838}\\{\scriptsize 0.852}\\{\scriptsize 0.818}} & \makecell[r]{0.836\\{\scriptsize 0.848}\\{\scriptsize 0.815}} & \makecell[r]{\textbf{0.839}\\{\scriptsize 0.852}\\{\scriptsize 0.819}} & \makecell[r]{0.825\\{\scriptsize 0.848}\\{\scriptsize 0.815}} \\
   & Semantic entropy & \makecell[r]{0.841\\{\scriptsize 0.854}\\{\scriptsize 0.824}} & \makecell[r]{\textbf{0.846}\\{\scriptsize 0.861}\\{\scriptsize 0.825}} & \makecell[r]{0.844\\{\scriptsize 0.854}\\{\scriptsize 0.823}} & \makecell[r]{\textbf{0.847}\\{\scriptsize 0.862}\\{\scriptsize 0.825}} & \makecell[r]{0.835\\{\scriptsize 0.854}\\{\scriptsize 0.823}} \\
\bottomrule
\end{tabular}
\end{adjustbox}
\caption{PRR on TriviaQA (validation split), LLM-judge supervision. \textbf{Bold} = significant improvement over baseline ($p < 0.05$, test set). \colorbox{yellow!40}{Highlight} = best PRR for model. }
\label{tab:prr_triviaqa_llm}
\end{table*}

\begin{table*}[t]
\centering
\footnotesize
\begin{adjustbox}{max width=\textwidth, max totalheight=0.92\textheight, keepaspectratio}
\setlength{\tabcolsep}{4pt}
\begin{tabular}{llrrrrr}
\toprule
\textbf{Model} & \textbf{Features} & \textbf{baseline} & \makecell{\textbf{latent}\\\textbf{gated}} & \makecell{\textbf{oracle}\\\textbf{gated}} & \makecell{\textbf{latent}\\\textbf{selective}} & \makecell{\textbf{oracle}\\\textbf{selective}} \\
\midrule
  \multirow{7}{*}{Llama-3.1-8B} & all UQ & \makecell[r]{0.582\\{\scriptsize 1.000}\\{\scriptsize 0.478}} & \cellcolor{yellow!40}\makecell[r]{\textbf{0.829}\\{\scriptsize 0.795}\\{\scriptsize 0.828}} & \makecell[r]{\textbf{0.703}\\{\scriptsize -0.480}\\{\scriptsize 0.864}} & \makecell[r]{\textbf{0.798}\\{\scriptsize 0.633}\\{\scriptsize 0.833}} & \makecell[r]{0.552\\{\scriptsize 0.795}\\{\scriptsize 0.647}} \\
   & CoCoA & \makecell[r]{0.513\\{\scriptsize 1.000}\\{\scriptsize 0.419}} & \makecell[r]{0.506\\{\scriptsize 0.542}\\{\scriptsize 0.476}} & \makecell[r]{\textbf{0.600}\\{\scriptsize 0.213}\\{\scriptsize 0.724}} & \makecell[r]{0.392\\{\scriptsize 0.334}\\{\scriptsize 0.395}} & \makecell[r]{0.498\\{\scriptsize -1.000}\\{\scriptsize 0.620}} \\
   & Semantic energy & \makecell[r]{0.462\\{\scriptsize -0.752}\\{\scriptsize 0.720}} & \makecell[r]{\textbf{0.496}\\{\scriptsize 0.076}\\{\scriptsize 0.600}} & \makecell[r]{0.597\\{\scriptsize 1.000}\\{\scriptsize 0.696}} & \makecell[r]{0.491\\{\scriptsize -0.480}\\{\scriptsize 0.756}} & \makecell[r]{\textbf{0.520}\\{\scriptsize -0.752}\\{\scriptsize 0.639}} \\
   & Semantic entropy & \makecell[r]{0.069\\{\scriptsize 0.633}\\{\scriptsize -0.023}} & \makecell[r]{0.573\\{\scriptsize 0.795}\\{\scriptsize 0.523}} & \makecell[r]{\textbf{0.386}\\{\scriptsize -0.480}\\{\scriptsize 0.468}} & \makecell[r]{0.566\\{\scriptsize 0.633}\\{\scriptsize 0.547}} & \makecell[r]{\textbf{0.430}\\{\scriptsize -0.480}\\{\scriptsize 0.523}} \\
\bottomrule
\end{tabular}
\end{adjustbox}
\caption{PRR on NCQA (test split), LLM-judge supervision. \textbf{Bold} = significant improvement over baseline ($p < 0.05$, test set). \colorbox{yellow!40}{Highlight} = best PRR for model.}
\label{tab:prr_conflictingqa_llm}
\end{table*}

\begin{table*}[t]
\centering
\footnotesize
\begin{adjustbox}{max width=\textwidth, max totalheight=0.92\textheight, keepaspectratio}
\setlength{\tabcolsep}{4pt}
\begin{tabular}{llrrrrr}
\toprule
\textbf{Model} & \textbf{Features} & \textbf{baseline} & \makecell{\textbf{latent}\\\textbf{gated}} & \makecell{\textbf{oracle}\\\textbf{gated}} & \makecell{\textbf{latent}\\\textbf{selective}} & \makecell{\textbf{oracle}\\\textbf{selective}} \\
\midrule
  \multirow{7}{*}{Llama-3.1-8B} & all UQ & \makecell[r]{0.799\\{\scriptsize 0.812}\\{\scriptsize 0.784}} & \makecell[r]{0.798\\{\scriptsize 0.810}\\{\scriptsize 0.783}} & \cellcolor{yellow!40}\makecell[r]{0.799\\{\scriptsize 0.812}\\{\scriptsize 0.783}} & \makecell[r]{0.797\\{\scriptsize 0.808}\\{\scriptsize 0.784}} & \makecell[r]{0.797\\{\scriptsize 0.810}\\{\scriptsize 0.783}} \\
   & CoCoA & \makecell[r]{0.768\\{\scriptsize 0.783}\\{\scriptsize 0.751}} & \makecell[r]{\textbf{0.771}\\{\scriptsize 0.786}\\{\scriptsize 0.754}} & \makecell[r]{0.769\\{\scriptsize 0.783}\\{\scriptsize 0.751}} & \makecell[r]{0.767\\{\scriptsize 0.783}\\{\scriptsize 0.750}} & \makecell[r]{0.768\\{\scriptsize 0.783}\\{\scriptsize 0.751}} \\
   & Semantic energy & \makecell[r]{0.752\\{\scriptsize 0.765}\\{\scriptsize 0.736}} & \makecell[r]{0.753\\{\scriptsize 0.766}\\{\scriptsize 0.737}} & \makecell[r]{0.753\\{\scriptsize 0.764}\\{\scriptsize 0.737}} & \makecell[r]{0.750\\{\scriptsize 0.764}\\{\scriptsize 0.736}} & \makecell[r]{0.752\\{\scriptsize 0.764}\\{\scriptsize 0.737}} \\
   & Semantic entropy & \makecell[r]{0.754\\{\scriptsize 0.769}\\{\scriptsize 0.736}} & \makecell[r]{\textbf{0.756}\\{\scriptsize 0.771}\\{\scriptsize 0.738}} & \makecell[r]{0.754\\{\scriptsize 0.768}\\{\scriptsize 0.736}} & \makecell[r]{0.754\\{\scriptsize 0.771}\\{\scriptsize 0.736}} & \makecell[r]{0.754\\{\scriptsize 0.768}\\{\scriptsize 0.736}} \\
\bottomrule
\end{tabular}
\end{adjustbox}
\caption{AUROC (error prediction) on AmbigQA (validation split), LLM-judge supervision. \textbf{Bold} = significant improvement over baseline ($p < 0.05$, test set). \colorbox{yellow!40}{Highlight} = best AUROC for model. }
\label{tab:auroc_ambigqa_llm}
\end{table*}

\begin{table*}[t]
\centering
\footnotesize
\begin{adjustbox}{max width=\textwidth, max totalheight=0.92\textheight, keepaspectratio}
\setlength{\tabcolsep}{4pt}
\begin{tabular}{llrrrrr}
\toprule
\textbf{Model} & \textbf{Features} & \textbf{baseline} & \makecell{\textbf{latent}\\\textbf{gated}} & \makecell{\textbf{oracle}\\\textbf{gated}} & \makecell{\textbf{latent}\\\textbf{selective}} & \makecell{\textbf{oracle}\\\textbf{selective}} \\
\midrule
  \multirow{7}{*}{Llama-3.1-8B} & all UQ & \makecell[r]{0.904\\{\scriptsize 0.914}\\{\scriptsize 0.891}} & \makecell[r]{0.904\\{\scriptsize 0.914}\\{\scriptsize 0.891}} & \makecell[r]{0.903\\{\scriptsize 0.913}\\{\scriptsize 0.890}} & \cellcolor{yellow!40}\makecell[r]{0.904\\{\scriptsize 0.914}\\{\scriptsize 0.891}} & \makecell[r]{0.903\\{\scriptsize 0.913}\\{\scriptsize 0.890}} \\
   & CoCoA & \makecell[r]{0.884\\{\scriptsize 0.893}\\{\scriptsize 0.872}} & \makecell[r]{\textbf{0.885}\\{\scriptsize 0.893}\\{\scriptsize 0.872}} & \makecell[r]{0.884\\{\scriptsize 0.892}\\{\scriptsize 0.871}} & \makecell[r]{\textbf{0.885}\\{\scriptsize 0.894}\\{\scriptsize 0.872}} & \makecell[r]{0.884\\{\scriptsize 0.892}\\{\scriptsize 0.871}} \\
   & Semantic energy & \makecell[r]{0.874\\{\scriptsize 0.885}\\{\scriptsize 0.858}} & \makecell[r]{\textbf{0.875}\\{\scriptsize 0.887}\\{\scriptsize 0.860}} & \makecell[r]{0.874\\{\scriptsize 0.885}\\{\scriptsize 0.858}} & \makecell[r]{\textbf{0.876}\\{\scriptsize 0.887}\\{\scriptsize 0.861}} & \makecell[r]{0.874\\{\scriptsize 0.885}\\{\scriptsize 0.858}} \\
   & Semantic entropy & \makecell[r]{0.880\\{\scriptsize 0.891}\\{\scriptsize 0.865}} & \makecell[r]{\textbf{0.881}\\{\scriptsize 0.893}\\{\scriptsize 0.866}} & \makecell[r]{0.880\\{\scriptsize 0.891}\\{\scriptsize 0.865}} & \makecell[r]{\textbf{0.882}\\{\scriptsize 0.893}\\{\scriptsize 0.867}} & \makecell[r]{0.880\\{\scriptsize 0.891}\\{\scriptsize 0.865}} \\
\bottomrule
\end{tabular}
\end{adjustbox}
\caption{AUROC (error prediction) on TriviaQA (validation split), LLM-judge supervision. \textbf{Bold} = significant improvement over baseline ($p < 0.05$, test set). \colorbox{yellow!40}{Highlight} = best AUROC for model. }
\label{tab:auroc_triviaqa_llm}
\end{table*}

\begin{table*}[t]
\centering
\footnotesize
\begin{adjustbox}{max width=\textwidth, max totalheight=0.92\textheight, keepaspectratio}
\setlength{\tabcolsep}{4pt}
\begin{tabular}{llrrrrr}
\toprule
\textbf{Model} & \textbf{Features} & \textbf{baseline} & \makecell{\textbf{latent}\\\textbf{gated}} & \makecell{\textbf{oracle}\\\textbf{gated}} & \makecell{\textbf{latent}\\\textbf{selective}} & \makecell{\textbf{oracle}\\\textbf{selective}} \\
\midrule
  \multirow{7}{*}{Llama-3.1-8B} & all UQ & \makecell[r]{0.718\\{\scriptsize 1.000}\\{\scriptsize 0.654}} & \cellcolor{yellow!40}\makecell[r]{0.846\\{\scriptsize 0.824}\\{\scriptsize 0.846}} & \makecell[r]{\textbf{0.747}\\{\scriptsize 0.235}\\{\scriptsize 0.875}} & \makecell[r]{\textbf{0.840}\\{\scriptsize 0.765}\\{\scriptsize 0.856}} & \makecell[r]{\textbf{0.741}\\{\scriptsize 0.824}\\{\scriptsize 0.734}} \\
   & CoCoA & \makecell[r]{0.676\\{\scriptsize 1.000}\\{\scriptsize 0.609}} & \makecell[r]{\textbf{0.684}\\{\scriptsize 0.647}\\{\scriptsize 0.679}} & \makecell[r]{\textbf{0.682}\\{\scriptsize 0.471}\\{\scriptsize 0.763}} & \makecell[r]{0.478\\{\scriptsize 0.647}\\{\scriptsize 0.449}} & \makecell[r]{0.558\\{\scriptsize 0.000}\\{\scriptsize 0.705}} \\
   & Semantic energy & \makecell[r]{0.665\\{\scriptsize 0.176}\\{\scriptsize 0.772}} & \makecell[r]{\textbf{0.672}\\{\scriptsize 0.412}\\{\scriptsize 0.744}} & \makecell[r]{0.701\\{\scriptsize 1.000}\\{\scriptsize 0.763}} & \makecell[r]{0.653\\{\scriptsize 0.118}\\{\scriptsize 0.776}} & \makecell[r]{0.598\\{\scriptsize 0.176}\\{\scriptsize 0.712}} \\
   & Semantic entropy & \makecell[r]{0.533\\{\scriptsize 0.706}\\{\scriptsize 0.519}} & \makecell[r]{0.716\\{\scriptsize 0.824}\\{\scriptsize 0.696}} & \makecell[r]{\textbf{0.592}\\{\scriptsize 0.235}\\{\scriptsize 0.638}} & \makecell[r]{0.722\\{\scriptsize 0.765}\\{\scriptsize 0.708}} & \makecell[r]{\textbf{0.541}\\{\scriptsize 0.235}\\{\scriptsize 0.647}} \\
\bottomrule
\end{tabular}
\end{adjustbox}
\caption{AUROC (error prediction) on NCQA (test split), LLM-judge supervision. \textbf{Bold} = significant improvement over baseline ($p < 0.05$, test set). \colorbox{yellow!40}{Highlight} = best AUROC for model.}
\label{tab:auroc_conflictingqa_llm}
\end{table*}

\end{document}